\documentclass[letterpaper, 10 pt, conference]{ieeeconf}  

\IEEEoverridecommandlockouts                              

\usepackage{amsmath,amssymb,amsfonts}
\usepackage{algorithmic}
\usepackage{graphicx}
\usepackage{braket}
\usepackage{textcomp}
\usepackage{xcolor}
\usepackage{subcaption}
\usepackage{float}
\usepackage{multicol}
\usepackage{caption}
\usepackage{soul}
\usepackage{multirow}
\usepackage{array}
\usepackage{makecell}
\usepackage{hyperref}

\usepackage[T1]{fontenc}
\usepackage[utf8]{inputenc}
\usepackage{babel}
\usepackage[font=small,labelfont=bf]{caption}

\usepackage[colorinlistoftodos]{todonotes}

\usepackage[normalem]{ulem}

\newcommand{\etal}{\textit{et al.}}

\title{\LARGE \bf
Grasp Stability Prediction with Sim-to-Real Transfer \\
from Tactile Sensing
}

\author{Zilin Si$^{1}$, Zirui Zhu$^{2}$,  Arpit Agarwal$^{1}$, Stuart Anderson$^{3}$ and Wenzhen Yuan$^{1}$
\thanks{$^{1}$Zilin Si, Arpit Agarwal and Wenzhen Yuan are with the Robotics Institute, Carnegie Mellon University
        {\tt\small \{zsi, arpita1, wenzheny\}@andrew.cmu.edu}}%
\thanks{$^{2}$Zirui Zhu is with the Department of Electrical  Engineering, Tsinghua University {\tt\small zhuzr17@mails.tsinghua.edu.cn}}%
\thanks{$^{3}$Stuart Anderson is with Meta Reality Labs Research {\tt\small stuarta@fb.com}}
}

\usepackage{array}
\makeatletter
\newcommand{\thickhline}{%
    \noalign {\ifnum 0=`}\fi \hrule height 1pt
    \futurelet \reserved@a \@xhline
}
\newcolumntype{"}{@{\hskip\tabcolsep\vrule width 1pt\hskip\tabcolsep}}
\makeatother

\begin{document}

\maketitle
\thispagestyle{empty}
\pagestyle{empty}

\begin{abstract}
Robot simulation has been an essential tool for data-driven manipulation tasks. However, most existing simulation frameworks lack either efficient and accurate models of physical interactions with tactile sensors or realistic tactile simulation. This makes the sim-to-real transfer for tactile-based manipulation tasks still challenging. In this work, we integrate simulation of robot dynamics and vision-based tactile sensors by modeling the physics of contact. This contact model uses simulated contact forces at the robot's end-effector to inform the generation of realistic tactile outputs. {To eliminate the sim-to-real transfer gap, we calibrate our physics simulator of robot dynamics, contact model, and tactile optical simulator with real-world data, and then} we demonstrate the effectiveness of our system on a zero-shot sim-to-real grasp stability prediction task where we achieve an average accuracy of 90.7\% on various objects. Experiments reveal the potential of applying our simulation framework to more complicated manipulation tasks. {We open-source our simulation framework at \href{https://github.com/CMURoboTouch/Taxim/tree/taxim-robot}{https://github.com/CMURoboTouch/Taxim/tree/taxim-robot}.}

\end{abstract}


\section{Introduction}
With the advent of deep learning, data-driven methods for solving various tasks like grasping~\cite{calandra2018more}, cloth manipulation~\cite{weng2022fabricflownet} and object properties estimation~\cite{yuan2017shape} have become ubiquitous in the robotic community. Data-driven techniques typically require large amounts of data generated from interaction with an environment, which incurs significant time and labor costs. In contrast to collecting data with real robots, simulation is an easy and effective tool for generating large-scale datasets. 

Dense vision-based tactile sensing provides rich contact information including contact shapes, textures, forces that can be used in various tasks such as shape reconstruction~\cite{wang20183d}, pose estimation~\cite{bauza2020tactile}, grasping~\cite{hogan2018tactile}, slip detection~\cite{7139016} etc. When solving these problems with data-driven approaches, simulating accurate contact dynamics and dense vision-based tactile readings can provide efficient data collection.  There exists multiple vision-based tactile simulation~\cite{agarwal2021simulation, gomes2021generation, hogan2021seeing,si2022taxim} which takes contact shapes as inputs and simulate the tactile images, and~\cite{wang2022tacto,church2022tactile} integrate the tactile simulation into robot dynamics simulation engines. However, these simulators are limited for dexterous manipulation tasks which also requires accurate contact dynamics simulation to handle contact forces, friction and the converting from contact physics to contact shapes of tactile sensors.~\cite{moisio2012simulation,9158822} model the contact dynamics for low-resolution tactile sensors but they are rather simpler than modeling for the vision-based tactile sensors and cannot be applied for most perception tasks. In order to benefit both perception and manipulation tasks, we proposal to integrate a realistic dense vision-based tactile sensor simulation with existing robot simulation engines by bridging with a contact dynamics simulation model. 
\begin{figure}
    \centering
    \includegraphics[width=0.95\linewidth]{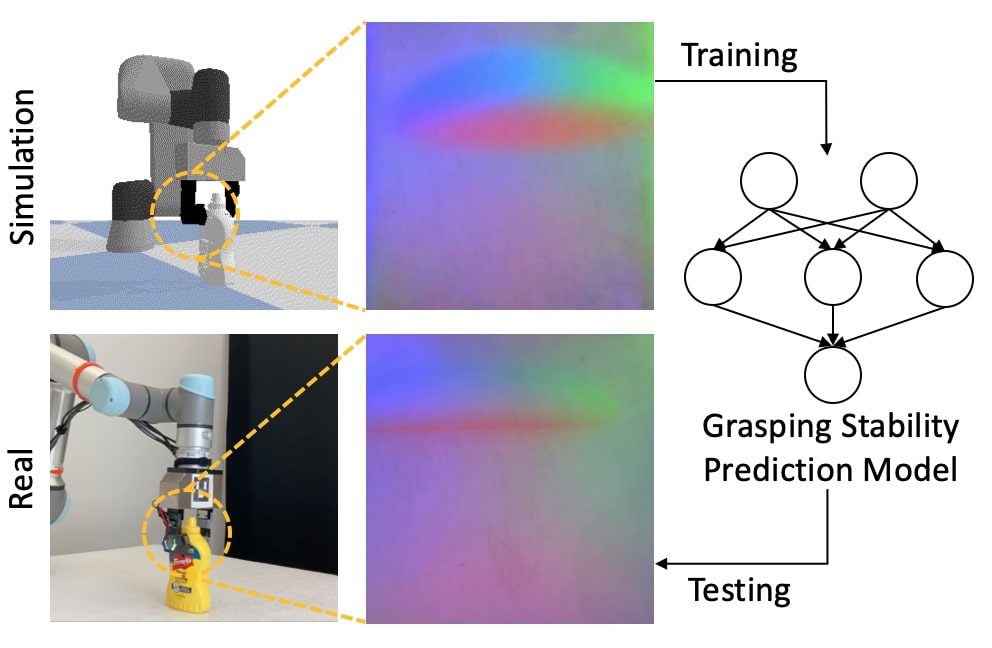}
    \caption{The pipeline of Sim2Real grasping stability prediction: we train an grasping stability prediction model using tactile images collected in our simulation and then test its performance on real-world tactile images.}
    \label{fig:teaser}
    \vspace{-5mm}
\end{figure}

The eventual goal of using simulation is to assist solving problems in real world, therefore sim-to-real transfer tasks make a good indicator to evaluate simulation frameworks. {Previous perception work with tactile sensing such as shape mapping~\cite{suresh2021efficient}, contact pose estimation~\cite{wang2022tacto} and edge/surface following~\cite{church2022tactile} only used simulation of tactile readings based on geometry and succeeded in sim-to-real transfer. However, sim-to-real grasping requires not only realistic geometric tactile simulation but also accurate contact dynamics simulation, where the latter one was not achieved in the previous simulation frameworks. To this end, we present the first integrated simulation framework with modeling the contact dynamics and combining it with a dense vision-based tactile simulator. By calibrating physics parameters of contact dynamics and tactile optical simulation with real-wolrd data, we show our simulation framework can successfully be applied to sim-to-real grasping stability prediction task with high accuracy.}

In this work, we model the contact dynamics to convert the contact forces to the contact deformation and then integrate the Taxim~\cite{si2022taxim}, a realistic vision-based tactile simulator, to improve the tactile simulation system’s fidelity. We apply our simulation framework on a sim-to-real grasp stability prediction task. We learn a model given tactile images to predict the grasp outcomes completely in simulation and then perform zero-shot sim-to-real transfer to test on tactile images from real grasp experiments. To the best of our knowledge, this is the first successful attempt at zero-shot sim-to-real transfer for grasp stability prediction with dense vision-based tactile sensing. This shows the potential of our simulation framework on more complex manipulation tasks.

\section{RELATED WORK}

\subsection{Grasping with tactile sensing}
Tactile sensing has been widely applied to grasping tasks given their rich contact information. Schill \etal~\cite{schill2012learning} learnt a classifer from 6 tactile sensors on a robot hand to continuously estimate the grasp stability during the grasp until reach the stable grasp. Bekiroglu \etal~\cite{bekiroglu2016probabilistic} learnt a latent variable probabilistic model from vision, tactile and action parameters. They used the conditional of this model to estimate the grasp stability. Calandra \etal~\cite{calandra2017feeling} showed that visuo-tactile deep neural network model can improve the ability to predict the grasp. And furthermore, the author proposed an action-conditional model to learn the regrasping policies from visual and tactile sensing in~\cite{calandra2018more}.
Note in all the above cases, a data collection stage in the real world was performed to build a learning model by grasping various objects. This process is time consuming and requires a human in the loop to reset the moving objects during grasping. 

Alternatively, to overcome the limitation in the data size, an exploration-based grasping method with visual and tactile data was developed and tested in the simulation~\cite{de2021simultaneous}. Bekiroglu \etal~\cite{bekiroglu2011assessing} learnt a probabilistic model of grasp stability, given tactile signal, joint configuration of the hand, object shape class and approach vector of the hand. They trained and tested the model on both simulated and real data but without sim-to-real transfer. Hogan \etal~\cite{hogan2018tactile} proposed a grasp quality metric based on tactile images and used the simulated regrasp candidates along with their tactile images to search for grasp adjustment. To predict the grasp stability given tactile images, our solution is to train completely in simulation, given a high fidelity simulation system {with calibrated physical parameters with real-world examples} and test directly on a real robot.

\subsection{Simulation of Tactile Sensors}
There are a lot of physics engines available to robotics practitioners~\cite{erez2015simulation} which allow full robot simulation with multiple sensing modalities. However, tactile simulation is limited in those simulators as deformable soft surfaces in tactile sensors are hard to simulate accurately and efficiently. Due to above, tactile simulation is still a challenging research area. Existing work~\cite{pezzementi2010characterization, narang2020interpreting, narang2021sim, sferrazza2019ground} build the mechanics models to simulate the soft body deformation for tactile sensors. For vision-based tactile sensors like GelSight~\cite{gelsight}, optical simulation~\cite{gomes2021generation, hogan2021seeing, agarwal2021simulation, si2022taxim} is also essential as it is used to measure shapes of objects in contact. {Several sim-to-real robot perception work~\cite{bauza2020tactile,suresh2021efficient, gao2022objectfolder} has revealed the accuracy and sufficiency of tactile optical simulation, and we leverage the existing tactile sensing simulation in robot simulator to make better use of them.}

To integrate the tactile simulation with physics simulation for manipulation tasks, Moisio \etal~\cite{moisio2012simulation} simulated
a low-resolution tactile sensor considering soft contacts and full friction description and applied it on grasping tasks. Kappassov \etal~\cite{9158822} presented a tactile simulation framework for tactile arrays in Gazebo simulation environment considering the effect of contact forces and showed tactile servoing applications. Wang \etal~\cite{wang2022tacto} presented TACTO, a simulator using pyrender to simulate optical tactile sensors and combined it to a physics simulator PyBullet. Compared to those simulation frameworks, we combine the simulation of contact physics including forces, frictions and slips with the vision-based tactile simulation of contact shapes, which allows the potential to simulate the more complicated grasping scenarios with slip and efficient sim-to-real transfer.

\subsection{Sim-to-Real Learning}
The control policies learned in simulation can be applied to real robots by framing the problem as transfer learning between the data distribution of simulation and real worlds. Dang \etal~\cite{dang2014stable} presented a learning approach to estimate the grasp stability and make hand adjustments based on low-resolution tactile sensing data from robot hand. They realized sim-to-real transfer but their testing objects are rather limited in number and have simpler shapes. In~\cite{church2022tactile}, the authors proposed an image-conditioned generator network that translates between real and simulated images. They used this network to transfer policies trained in simulation for various tactile manipulation tasks. In~\cite{wu2019mat}, the authors set up two multi-fingered hands with tactile sensors in PyBullet, used it to learn a reinforcement learning policy for grasping and transferred the learnt policy to real world environments.
Mahler \etal~\cite{mahler2017learning} considered the problem of bin picking by simulating quasi-static physics in PyBullet with a parallel-jaw gripper. They posed the sim-to-real task as a transferring GQ-CNN features between simulation and real world data from Dex-Net 2.0 dataset~\cite{mahler2017dex}. We train a grasp stability prediction model based on dense vision-based tactile images in our simulation framework and show that the model can be successfully transferred to the real data without any explicitly transfer step.  

\begin{figure}[t]
    \centering
    \includegraphics[width=0.98\linewidth]{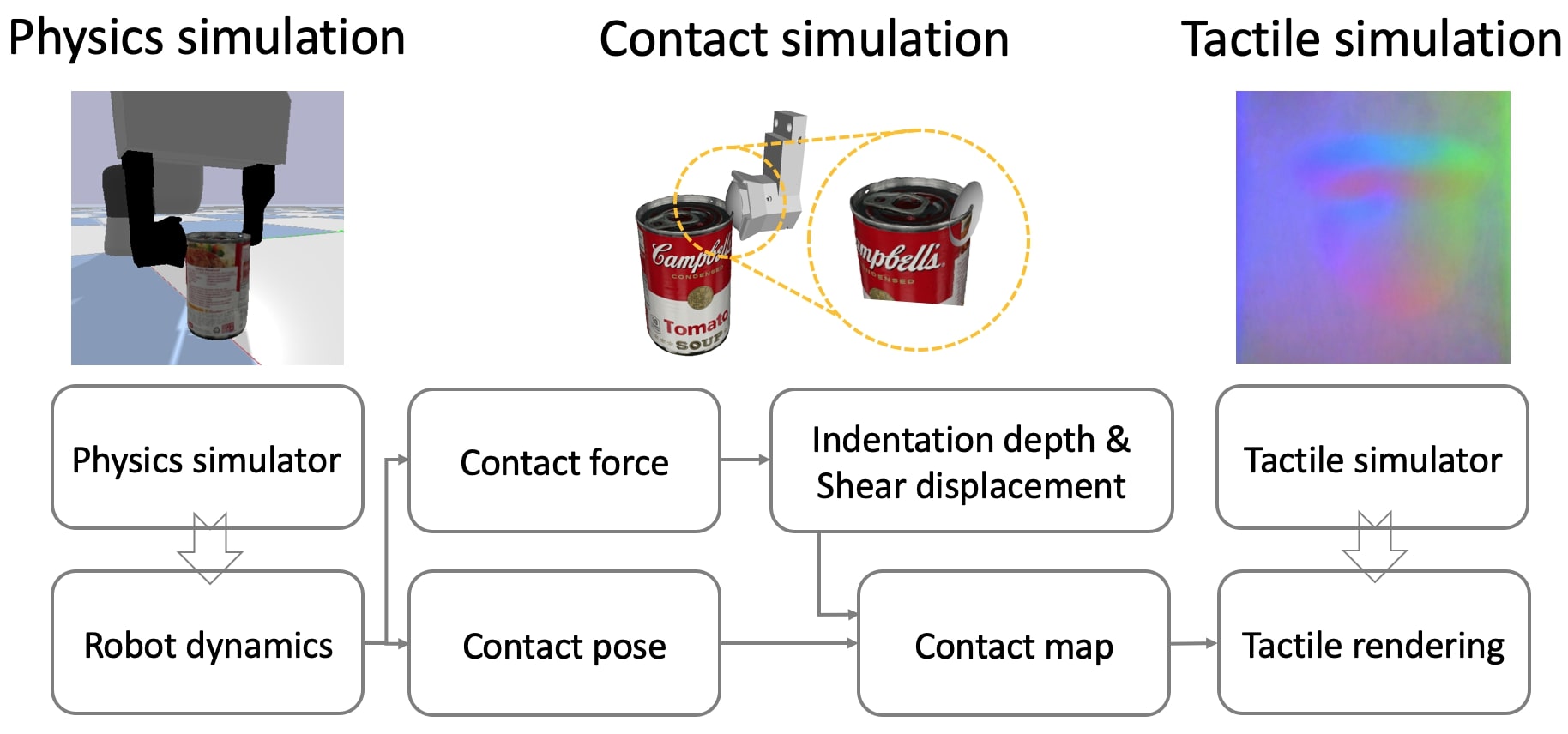}
    \caption{Our proposed simulation framework includes physics simulation, contact simulation and tactile simulation. Physics simulator handles the robot dynamics which provide the contact forces and poses. Contact models map them to indentation depths and shear displacements of the contact, and generate the contact map to feed into the tactile simulator. Tactile simulation renders the RGB tactile images.}
    \label{fig:simframe}
    \vspace{-5mm}
\end{figure}
\section{Simulation Framework}
In this section, we present our integrated simulation framework with tactile sensing. The framework includes three parts as shown in Fig.~\ref{fig:simframe}: physics simulation in Section~\ref{sec::pgysics_sim}, contact simulation in Section~\ref{sec::contact_sim} and tactile simulation in Section~\ref{sec::tactile_sim}. We use PyBullet to simulate the physics, and transfer the contact forces and poses to the contact deformation of a GelSight tactile sensor~\cite{gelsight}, which has a soft surface to interact with the object and an embedded camera to convert the contact geometries to RGB images. Then the tactile simulator renders tactile images according to the contact deformation.

\subsection{Physics Simulation}\label{sec::pgysics_sim}
Physics simulation reproduces the real world dynamic interaction between the robots and objects, and its accuracy directly impacts sim-to-real transfer. We use PyBullet as our physics simulation engine. For the grasping task, we load the robot arm, gripper and a GelSight mounted on the gripper with proper geometries and links as shown in Fig.~\ref{fig:teaser}.

Grasping requires accurate contact simulation, therefore physics parameters setting such as friction, objects' mass and their center of mass becomes essential. In order to match the simulation with the real grasping scenario, we measure the weights and center of mass of objects in reality as the reference. We then estimate the friction of the contact surface by calibrating with the real world data. Specifically, we search for the best friction coefficient by adjusting friction values in simulation to minimize the grasping label mismatching between real and simulated data under the same grasping configurations. {See Section~\ref{sec:frictionOpt} for more details.}

\subsection{Contact Simulation}\label{sec::contact_sim}
Contact simulation refers to simulating the deformation of GelSight's soft surface under applied contact forces during the interaction with the object. 
PyBullet simulates the robot dynamics, but it lacks accurate contact model since the soft body deformation of tactile sensors is not able to be simulated with only rigid body collision.
Instead of applying computationally costly soft body simulation, we use an simplified model that maps contact forces to the deformation of the tactile sensor.

\begin{figure}[t]
    \centering
    \includegraphics[width=0.98\linewidth]{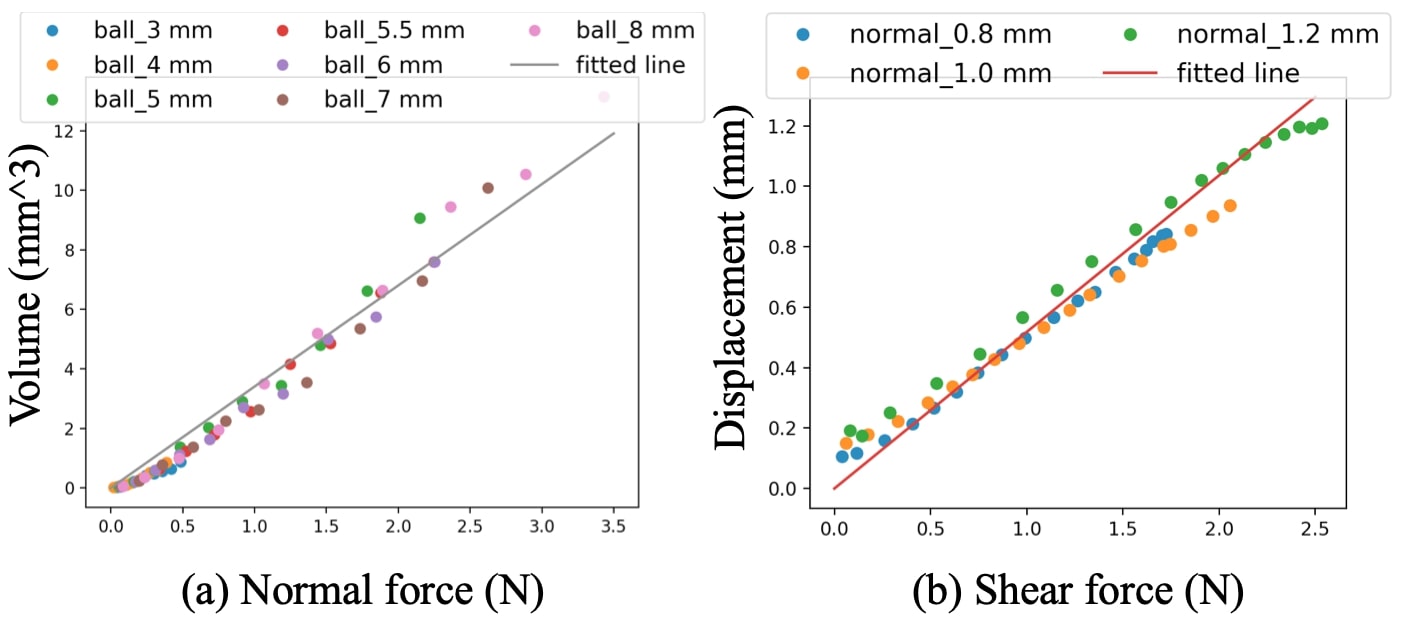}
    \caption{Experiment results showed that the contact force is almost linear to the deformation of the GelSight in both normal and shear directions, regardless of the contact geometry. (a) Experiment results of the contact volume vs. normal force for balls with different diameters. (b) Experiment results of the sum of markers' shear motion vs. overall shear force.
    }
    \label{fig:contact_mapping}
    \vspace{-5mm}
\end{figure}

When the tactile sensor touches the object, PyBullet calculates the contact force and relative poses between the object and the sensor. To simulate the object's indentation into the tactile sensor's surface under the normal loading and sliding on the contact surface under the shear loading, we map the contact force to the indentation depth and the shear motion of the contact. {Enlighten by~\cite{gelsight} and from experimental measurements on our real GelSight sensor plotted in Fig.~\ref{fig:contact_mapping}, we approximate the linear mapping between the normal force and the indentation volume, and between the shear force and the shear motion of the contact. We characterize their coefficient $k_n$ and $k_s$ as:}
\begin{equation}
\begin{split}
    & V = k_n F_n \\
    & D = k_s F_s
\end{split}
\end{equation}
where $V$ is the indentation volume, $F_n$ is the normal force; $D$ is the shear displacement, $F_s$ is the shear force. 

Recovering the indentation depth map from the volume $V$ does not have a close-form solution,
therefore we use the binary searching to find a best estimated depth map. Given an initial depth map, we integrate the indentation depth $d(A)$ within the contact area $A$ to the volume as:
\begin{equation}
    V_{est} = \int d(A) dA
\end{equation}
We iteratively adjust the $d(A)$ to minimize the error between the estimated volume $V_{est}$ and the target volume $V$ to find the best solution.

To simulate the contact shape, we utilize the PyRender in a similar way as TACTO~\cite{wang2022tacto} and place an virtual depth camera behind the sensor surface to capture the 3D contact shape, defined as the \textit{contact map}. Given the indentation depth, shear displacement and the relative poses between the object and the tactile sensor, we move the object along the normal direction of contact with indentation depth and the tangential direction with the shear displacement in PyRender and then render the \textit{contact map}. 

Note that PyBullet reserves a collision margin between two colliding objects to ensure numerical stability where two objects considered as collided are still separated by a distance from each other. This margin even changes when the object has non-convex geometric surfaces.
Therefore, we adapt this margin in PyRender to get precise contact indentation depth.

\subsection{Tactile Simulation}\label{sec::tactile_sim}
After getting the contact map from the contact model, we render the GelSight tactile images with an state-of-the-art simulation model, Taxim~\cite{si2022taxim}. Taxim uses a lookup table to map the contact shapes to the tactile images. The lookup table is calibrated with a real sensor used in the experiment.

\begin{figure*}[ht]
\centering
\begin{minipage}[t]{.72\textwidth}
\includegraphics[width=0.98\textwidth]{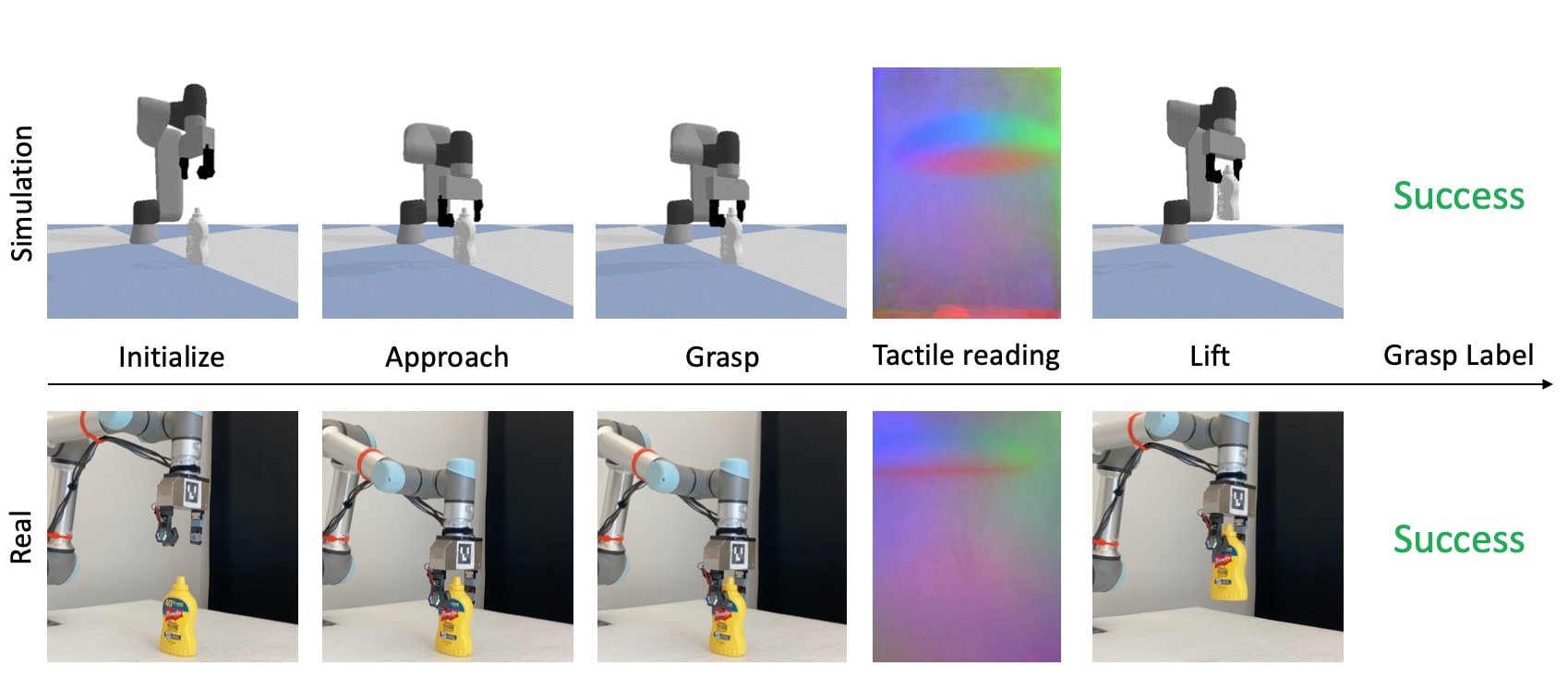}
\caption{Grasp pipeline for both simulation and real experiments. We initialize the robot on top of the object, move the gripper down to a preset height, close the gripper with a preset grasping force to grasp the object, and then lift it. We record the tactile readings from a GelSight sensor after grasping.}
\label{fig:grasp_pipeline}
\end{minipage}\qquad
\begin{minipage}[t]{.22\textwidth}
\includegraphics[width=0.98\textwidth]{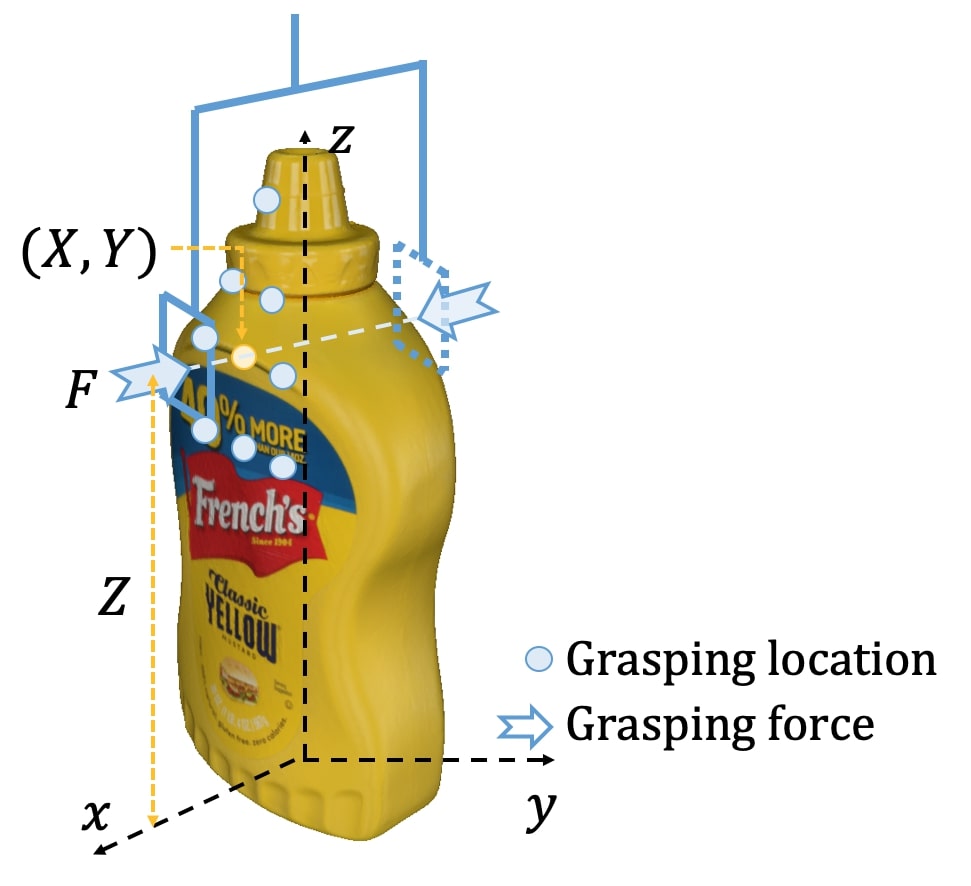}
\caption{Demonstration of a grasping configuration: it is defined as the grasp location, height and force.
}
\label{fig:grasp_config}
\vspace{-5mm}
\end{minipage}
\end{figure*}

\section{Sim-to-Real Grasping Prediction}
We formalize the grasping stability prediction as a supervise learning problem where we use the tactile images during grasping to predict the grasp outcomes.
We use simulated tactile images and labels from our simulation framework to train the learning model and test it on real-world data with a zero-shot sim-to-real transfer.

\begin{figure}
    \centering
    \includegraphics[width=0.98\linewidth]{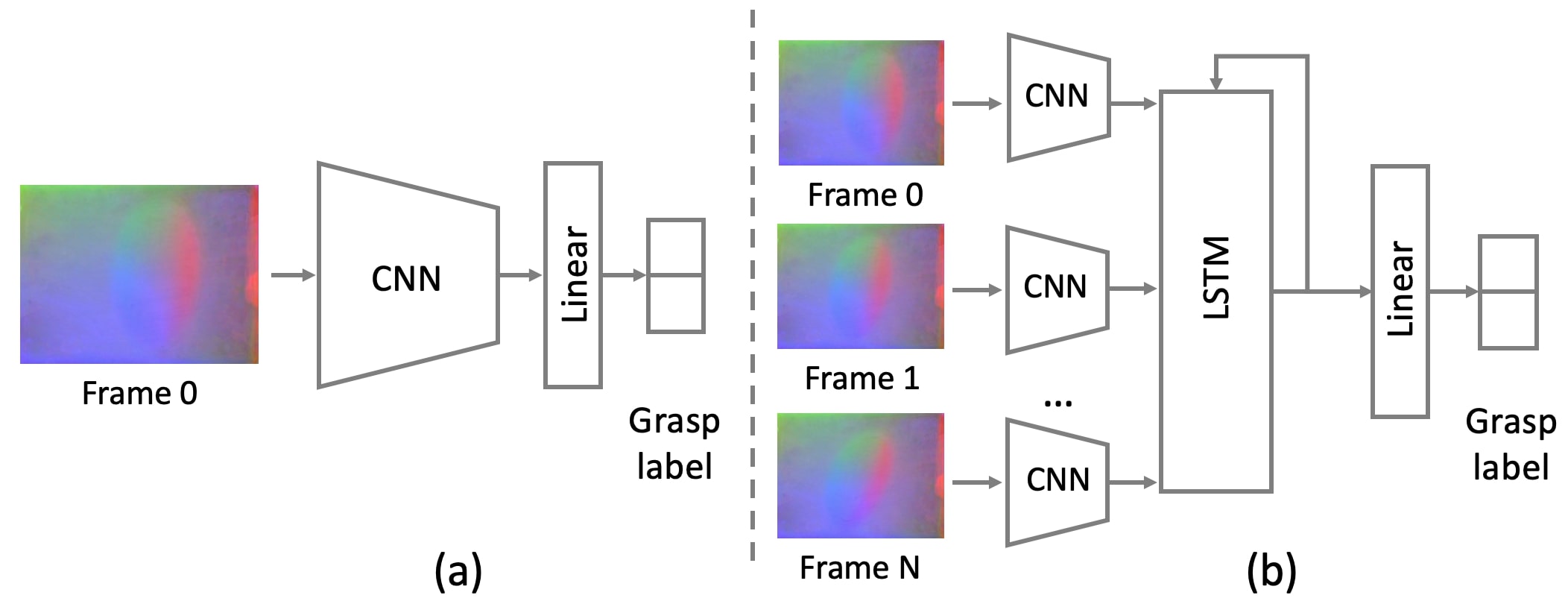}
    \caption{Grasp stability prediction networks. (a) We input single tactile images to a feature extractor (CNN) and a classifier (MLP) to predict the grasp results. (b) We input a sequence of tactile images to feature extractors (CNN), a LSTM module and then a classifier (MLP) to predict the grasp results.}
    \label{fig:network}
    \vspace{-5mm}
\end{figure}

\subsection{Grasp Stability Prediction Model}
We try to classify the grasp stability into binary labels: success when the object can be stably lifted and failure otherwise.
We predict grasp outcomes based on a single or sequential tactile images from a GelSight sensor. We take the tactile image at the moment of the object is grasped as the single input. To get the sequential input, we record tactile images for three seconds after the robot closes the gripper.

We build a neural network model to extract latent features of tactile images and then use a classifier to predict the grasp stability base on these features.
For the model with single tactile inputs, we use a pre-trained ResNet-18~\cite{he2015deep} as our feature extractor and a multi-layer perceptual (MLP) as classifier as shown in Fig.~\ref{fig:network} (a). For the sequential inputs, we feed the sequence of image features to a long short-term memory (LSTM) module and then forward the last hidden state's output to the classifier as shown in Fig.~\ref{fig:network} (b).


\subsection{Grasping Pipeline}
\begin{figure}[t]
    \centering
    \includegraphics[width=0.9\linewidth]{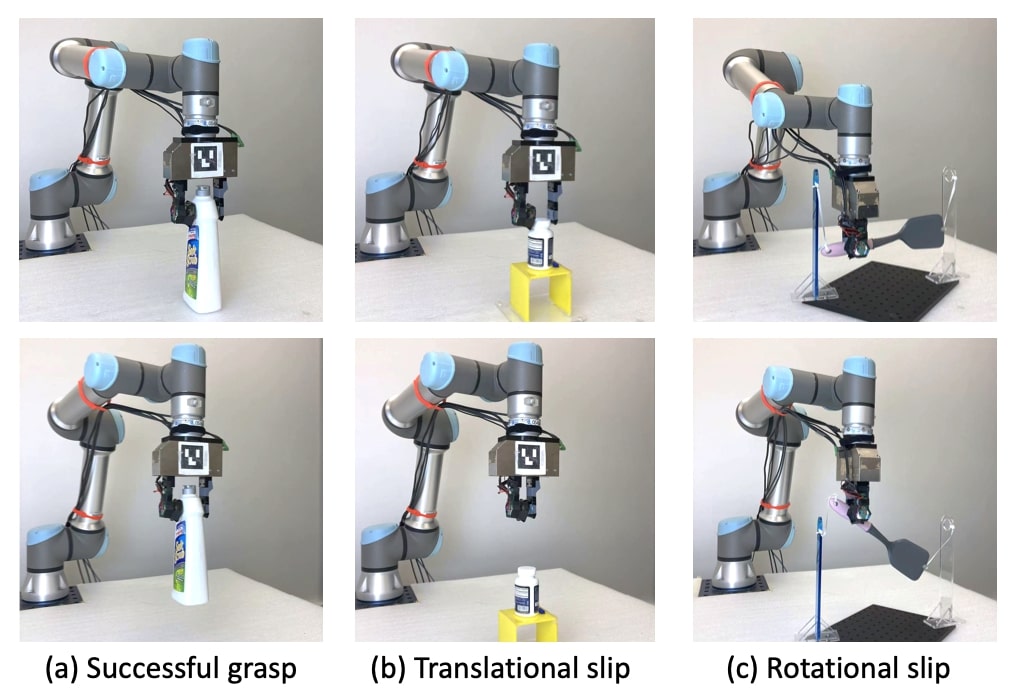}
    \caption{We classify grasp as successful grasp (a) and failed grasp (b), (c) including translational and rotational slip.
    }
    \label{fig:grasp_label}
    \vspace{-5mm}
\end{figure}

Our grasping pipeline includes initialization, approaching, grasping, lifting and labeling as shown in Fig~\ref{fig:grasp_pipeline}. We initialize the robot and the gripper on the top of the object. Given the specified grasping configuration, the robot rotates to the target orientation, moves straightly down to the grasping height, and adjust the grasping location on the plane parallel to table.
During the grasping, the gripper closes with a certain speed and force. Then till the gripper closes entirely, we record tactile images from the GelSight and use them as our grasping stability prediction model inputs. The robot then lifts the object for 18 cm. During the lifting, if the object remains stable in the gripper, we label the grasp as `success'; otherwise, if the object falls or slips in the gripper, we label the grasp as `failure'. There are two different kinds of failure: translational slip or rotational slip coming from lacking grasping forces or wrong grasping locations respectively as shown in Fig.~\ref{fig:grasp_label}. In experiments, we set certain thresholds, 15 cm translational movement and 0.1 radians rotational movement of objects to detect grasp failure.

\begin{figure*}[t]
    \centering
    \includegraphics[width=0.8\textwidth]{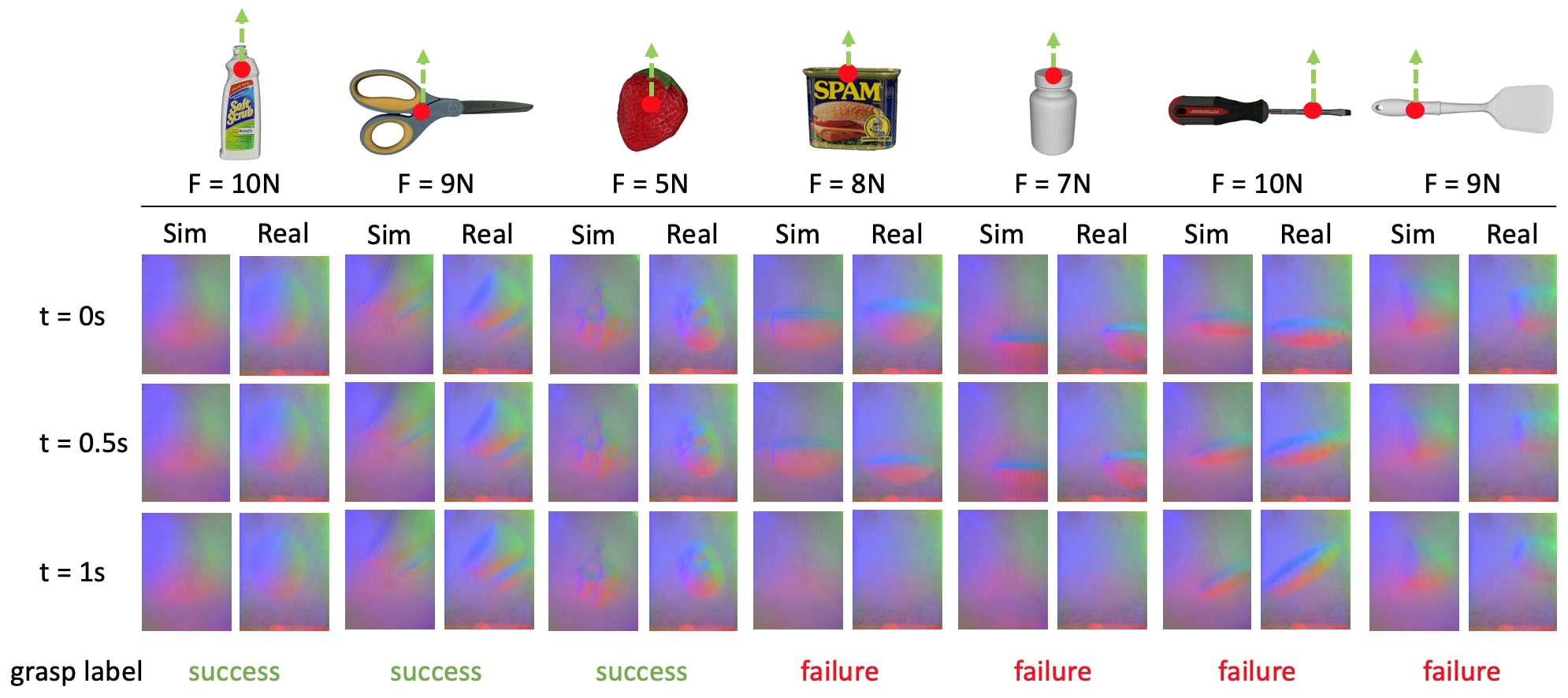}
    \caption{Examples of tactile readings under different grasping scenarios. Different grasping locations as marked on the object and grasping forces F lead to different grasping outcomes. We show the sequence of the tactile readings during grasping, where geometries of contact can be used to predict the grasping outcomes.
    }
    \label{fig:example}
\end{figure*}
\subsection{Grasping Configuration Generation}
We collect grasp data with various grasping configurations regarding the grasp location and force. We describe the grasping configuration as a vector $(F, X, Y, Z)$, where $F$ is the grasping force, $X,Y$ is grasping location on the horizontal plane, and $Z$ is the grasping height as shown in Fig.~\ref{fig:grasp_config}. We define the range of each dimension of grasping, discretize the grasping space and then conduct the grasping.

\section{EXPERIMENTS}
We conduct both simulated and real grasp experiments and evaluate our simulation framework's ability for sim-to-real transfer based on collected data.
Our data collection includes variance in three parameters of grasping: the target object, the grasping location, and the grasping force. We show the grasp stability prediction model's performance on both single and sequential tactile images. We provide details of our experiments in the following sections.

\subsection{Data Collection}
We use a UR5e robot with a Weiss WSG-50 gripper, and mount a GelSight~\cite{dong2017improved} sensor on one side of the gripper. We use the objects from YCB~\cite{Calli_2015} and GoogleScan~\cite{googleScannedObjects} datasets for grasping where the mesh models are available.
We conduct the same grasping process in both simulation and real as shown in Fig.~\ref{fig:grasp_pipeline}.

We selected twelve objects for both simulation and real-world data collection as shown in Table~\ref{Table:overall}. To demonstrate our model's ability to make predictions over a broad set of objects, we chose objects that provide a range of shapes, surface frictions, masses, and center of mass locations.

\textbf{Testing dataset from real world data} For each object, we grasp it at three to five different heights, ranging from the top of the object to the minimum reachable height; we vary the grasping location on the XY plane with three to six values depending on the size of the object; and we linearize the grasping force with six different values ranging from 5N to 10N. We also add mass (water or clay) ranging from 100g to 500g to light objects such as empty bottles to get more grasping failures.
These parameters are chosen so that the collected grasp data is approximately balanced with 363 successful and 389 failed grasps in total.

\textbf{Training dataset from simulated data} We use the same set of objects in simulation. For each object, we collected 100 to 150 grasp trials for training and 50 grasp trails for testing following the same configurations mentioned in testing dataset. For each grasp, we record the rendered tactile images and automatically label its corresponding grasp outcomes based on the pose changing of the object.

Some data examples are shown in Fig.~\ref{fig:example} where we mark the grasp locations, grasp forces, grasp labels and the corresponding tactile readings.

\begin{figure}[b]
    \centering
    \includegraphics[width=1.0\linewidth]{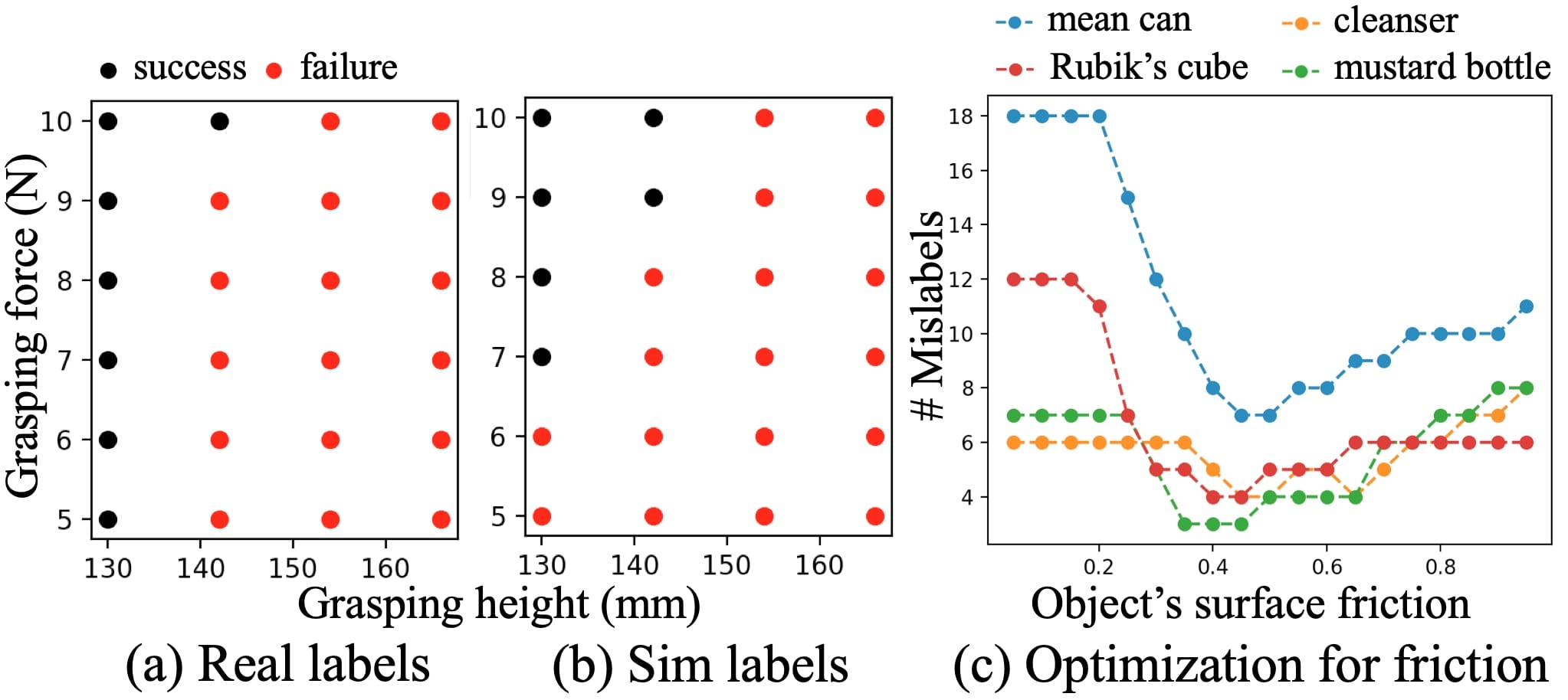}
    \caption{We optimize the friction coefficient of object surface by matching the grasping labels between simulated and real data under the same configuration of grasping heights and forces as shown in (a) and (b). We evaluate possible friction coefficients for several objects and choose the friciton values that minimize mislabeling as shown in (c).
    }
    \label{fig:friction}
    \vspace{-5mm}
\end{figure}

\begin{table*}[ht]
\centering
\setlength{\tabcolsep}{4pt}
\renewcommand{\arraystretch}{1.2}
\begin{tabular}{lcccccccccccccc}
\Xhline{2\arrayrulewidth}
{Object} & {\includegraphics[height=.05\textwidth]{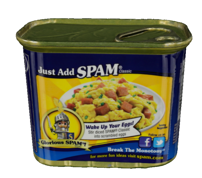}} & 
{\includegraphics[height=.05\textwidth]{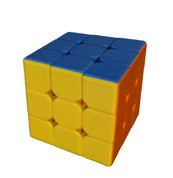}} &
{\includegraphics[height=.05\textwidth]{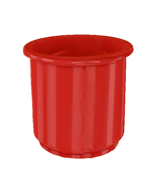}} & 
{\includegraphics[height=.05\textwidth]{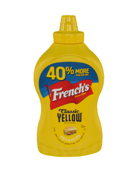}} &
{\includegraphics[height=.05\textwidth]{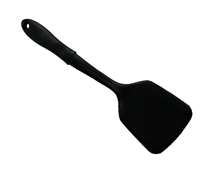}} & 
{\includegraphics[height=.05\textwidth]{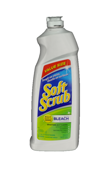}} &
{\includegraphics[height=.05\textwidth]{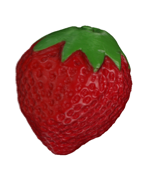}} &
{\includegraphics[height=.05\textwidth]{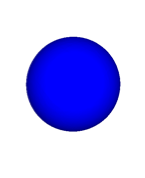}} &
{\includegraphics[height=.05\textwidth]{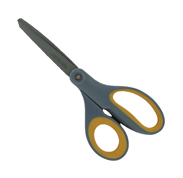}} & 
{\includegraphics[height=.05\textwidth]{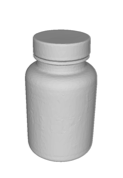}} &
{\includegraphics[height=.05\textwidth]{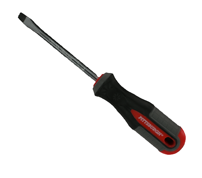}} &
{\includegraphics[height=.05\textwidth]{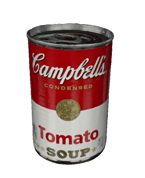}} &
{average}
 \\ \hline
Sim2Sim (TACTO~\cite{wang2022tacto}) & 0.860 & 0.900 & 0.960 & 0.860 & 0.720 & 0.940 & 1.000 & 0.850 & 0.920 & 0.960 & 0.550 & 0.650 & 0.841\\
Sim2Sim (Single) & 1.000 & 0.958 & 0.875 & 1.000  & 0.937  & 1.000 & 0.875 & 1.000 & 1.000 & 1.000 & 1.000 & 1.000 & 0.968\\ 
Sim2Sim (Sequential) & 1.000 & 1.000 & 1.000 & 1.000  & 1.000  & 1.000 & 1.000 & 1.000 & 1.000  & 1.000 & 1.000 & 1.000 & 1.000\\
\hline
Sim2Real (TACTO~\cite{wang2022tacto}) & 0.542 & 0.667 & 0.094 & 0.688 & 0.208 & 0.604 & 0.333 & 0.484 & 0.625 & 0.375 & 0.344 & 0.505 & 0.426\\
Sim2Real (Single) &  0.958   & 0.895  & 0.875 &  0.812 & 0.791 & 0.833 & 0.750 & 0.953 & 0.958  & 1.000 & 0.843 & 0.825 & 0.864\\ 
Sim2Real (Sequential) &  1.000   & 0.916 & 0.968 & 0.821  & 0.833 & 0.812 & 1.000 & 0.937 & 0.958 & 1.000 & 0.875 & 0.870 & 0.907\\
\Xhline{2\arrayrulewidth}
\end{tabular}
\caption{The result of grasp stability prediction. We test the prediction accuracy for both sim-to-sim and sim-to-real transfers with a single tactile image and sequences of tactile images. We compare the performance with TACTO~\cite{wang2022tacto}.}
\label{Table:overall}
\end{table*}

\subsection{Optimization of Friction} \label{sec:frictionOpt}
To eliminate the gap of sim-to-real transfer, we use a few real-world grasping examples (around 20) {apart from the testing dataset} for each object as reference, and tune the friction coefficient in simulation with an optimization process. As shown in Fig.~\ref{fig:friction}, for each object (here we use the mustard bottle as example), we generate the distribution of the grasping outcomes based on the grasping heights and forces for both simulated and real data. Adjusting the friction coefficient in simulation will lead to different distributions, and we search for the best friction coefficient based on how well the generated outcome distribution matches the real-world one. We discretize the friction coefficient search space in the range [0,1] with a step size of 0.05, then we plot the number of mislabeling in Fig.~\ref{fig:friction} (c). For each object, there is a point with the fewest mislabels and we denote the corresponding friction coefficient as the best one to use.

\subsection{Learning Model and Training Settings}
We use Res-Net 18~\cite{he2015deep} pre-trained on ImageNet as the tactile feature extractor. For single tactile inputs, we use a two-layer MLP 512-256-2 with ReLu activation and a 0.2 dropout after the first linear layer as classifier. For sequential tactile inputs, we use a three-layer LSTM module with 128 hidden layer size, then a two-layer MLP 128-64-2 with ReLu activation and a 0.2 dropout after the first linear layer as the classifier. 
For both learning models, we use the Adam optimizer with $1\times 10^{-4}$ learning rate, $5\times 10^{-5}$ weight decay and a ReduceLROnPlateau scheduler. We train the model with 8 batch size and 30 epochs. We set the learning rate of the feature extractor as 0.8 of the learning rate of the rest modules to prevent overfitting. We train individual prediction model for each object.

\subsection{Grasp Stability Prediction with Single Tactile Image}
We first test our sim-to-real grasp stability prediction model with single tactile images. The results are reported in Table~\ref{Table:overall} indicated as Sim2Sim (Single) and Sim2Real (Single). We denote the case of training on simulated data and testing on simulated data as ``Sim2Sim" and the case of training on simulated data and testing on real data as ``Sim2Real". From the table, we show our Sim2Sim prediction accuracy is 100\% excluding objects (Rubik's cube, cup, spatula, strawberry) which have ambiguous shapes. For those shapes, it is hard to locate the grasp based on single tactile images since they look similar or even the same.

We also show our Sim2Real prediction accuracy are all above 80\% except objects spatula and strawberry. The mesh models of these objects are too coarse compared to real objects, which leads to significant differences between simulated and real tactile images. On average, our Sim2Real gap is less than 10\% comparing to our Sim2Sim results. 

We conduct the same grasping experiments using TACTO~\cite{wang2022tacto} as a baseline and post results in the Table~\ref{Table:overall}. The results show that our method outperforms TACTO. A major failure of TACTO is caused by inaccurate contact model: it models the contact that the normal force is linear to the indentation depth, which does not match the physics of the sensor in the real world. In addition, TACTO does not properly calculate the collision margin so that it causes occasional failure in generating tactile signals upon contact.

\subsection{Grasp Stability Prediction with Sequential Tactile Image}
Single tactile images are less informative when objects have ambiguous geometries such as the flat surface of a Rubik's cube. Therefore we also test our simulation performance on models with sequential tactile images as inputs. Sim2Sim and Sim2Real accuracy results are shown in Table~\ref{Table:overall} indicated as Sim2Sim (Sequential) and Sim2Real (Sequential). Comparing to the single tactile results, we notice that the Sim2Real performance on objects Rubik's cube, cup, spatula, and strawberry is improved. And the average performance also improves to 90.7\%. This is mostly because the sequential data shows either the stable or the changing contacts from tactile images after lifting which can indicate the object's motion. And it serves as better features to predict the grasp outcomes.

\subsection{Effects of Dataset Size, Friction and Center of Mass}
We do ablation study on different setting of dataset size, friction on potted meat can object, and center of mass on scissor object in simulation. As shown in Table~\ref{tab:ablation}, the performance of Sim2Sim and Sim2Real both increase along with the increasing dataset size but reach the plateau after 200. This suggests that choosing dataset size between 100 to 200 is sufficient for this task. Comparing to the best choice 0.45 for friction coefficient and -0.03 for center of mass, even though the Sim2Sim accuracy stays high, the Sim2Real accuracy drops quickly with inaccurate parameters. This indicates that the physics settings affect the results significantly and it is necessary to set the proper physics parameters for effective sim-to-real transfer.

\begin{minipage}{0.9\linewidth}
  \begin{minipage}[b]{0.2\linewidth}
    \centering
    \includegraphics[width=0.95\textwidth]{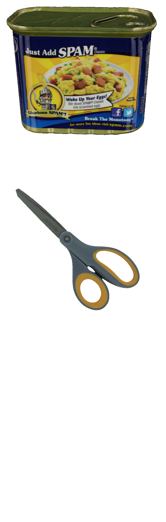}
  \end{minipage}
  \hfill
  \begin{minipage}[b]{0.78\linewidth}
    \centering
\scalebox{0.8}{\begin{tabular}{lllll}
\Xhline{2\arrayrulewidth}
\textbf{Dataset size}   & 50 & 100 & 200 & 500 \\
Sim2Sim Acc & 0.875 & 0.937 & 1.000 & 1.000  \\
Sim2Real Acc &  0.791  & 0.833 & 0.958 & 0.958 \\
\hline
\textbf{Friction coefficient}   & 0.2 & \textbf{0.45} & 0.8 &  \\
Sim2Sim Acc & 1.000 & 1.000 & 0.937 &  \\
Sim2Real Acc &  0.666  & 0.958 & 0.541 & \\
\hline
\textbf{Center of Mass}  & \textbf{-0.03} & 0.00 & 0.03 & \\
Sim2Sim Acc & 1.000 & 1.000 &  1.000  & \\
Sim2Real Acc &  0.958  & 0.708 & 0.625 & \\
\Xhline{2\arrayrulewidth}
\end{tabular}}
\label{tab:ablation}
\captionof{table}{Ablation study of dataset size, friction coefficient on potted meat can, and center of mass on scissor. 0.45 and -0.03 are the best parameters for friction and center of mass we used in our experiments.
}
\end{minipage}
\end{minipage}
\section{CONCLUSIONS}
In this work, we present a robot simulation framework with tactile sensing, where we integrate a contact model mapping the contact forces to the deformation of the soft surface of the GelSight tactile sensor.
We test the performance of our framework on grasp stability prediction task where we directly transfer the model trained on simulated data to real-world data. We show that our model can achieve 75\%-100\% accuracy in sim-to-real transfer on objects with various shapes and physical properties while hasn't been exposed to a single real tactile image. 

In the future, we would like to extend this work to sim-to-real policy transfer of grasp-regrasp planning based on our grasp prediction. In addition, we would like to combine tactile sensing with other modalities such as vision in simulation and explore its sim-to-real transfer. We would also like to apply our simulation framework on various manipulation tasks such as picking-and-placing, pushing, and sliding tasks. 


\addtolength{\textheight}{0 cm}   


\section*{ACKNOWLEDGMENT}
This work was funded by Meta AI research. The authors would like to thank Hung-Jui Huang, Yifan You, Dr. Roberto Calandra for the help and discussion on this work.

\bibliographystyle{bib/IEEEtran}
\bibliography{bib/IEEEabrv, bib/ref}

\end{document}